\documentclass[10pt,twocolumn,letterpaper]{article}

\usepackage{cvpr}              

\usepackage{amssymb}
\usepackage{pifont}
\usepackage[accsupp]{axessibility}  
\usepackage[normalem]{ulem}
\newcommand{\cmark}{\ding{51}} 
\newcommand{\xmark}{\ding{55}}

\definecolor{cvprblue}{rgb}{0.21,0.49,0.74}
\usepackage[pagebackref,breaklinks,colorlinks,allcolors=cvprblue]{hyperref}

\def\paperID{66} 
\def\confName{EarthVision}
\def\confYear{2026}

\title{Conflated Inverse Modeling to Generate Diverse and Temperature-Change Inducing Urban Vegetation Patterns}

\author{
Baris Sarper Tezcan$^{1}$ \quad
Hrishikesh Viswanath$^{1}$ \quad
Rubab Saher$^{2}$ \quad
Daniel Aliaga$^{1}$\\
$^{1}$Computer Science \qquad
$^{2}$Forestry and Natural Resources\\
Purdue University, West Lafayette, IN, USA\\
{\tt\small \{tezcanb,hviswan,rsaher,aliaga\}@purdue.edu}
}

\begin{document}
\maketitle
\begin{abstract}
Urban areas are increasingly vulnerable to thermal extremes driven by rapid urbanization and climate change. Traditionally, thermal extremes have been monitored using Earth-observing satellites and numerical modeling frameworks. For example, land surface temperature derived from Landsat or Sentinel imagery is commonly used to characterize surface heating patterns. These approaches operate as forward models, translating radiative observations or modeled boundary conditions into estimates of surface thermal states. While forward models can predict land surface temperature from vegetation and urban form, the inverse problem—determining spatial vegetation configurations that achieve a desired regional temperature shift—remains largely unexplored. This task is inherently underdetermined, as multiple spatial vegetation patterns can yield similar aggregated temperature responses. Conventional regression and deterministic neural networks fail to capture this ambiguity and often produce averaged solutions, particularly under data-scarce conditions. We propose a conflated inverse modeling framework that combines a predictive forward model with a diffusion-based generative inverse model to produce diverse, physically plausible image-based vegetation patterns conditioned on specific temperature goals. Our framework maintains control over thermal outcomes while enabling diverse spatial vegetation configurations—even when such combinations are absent from training data. Altogether, this work introduces a controllable inverse modeling approach for urban climate adaptation that accounts for the inherent diversity of the problem. Code is available at \href{https://github.com/barissarpertezcan/urban-vegetation-inverse-modeling/tree/main}{GitHub repository}.
\end{abstract}
    
\section{Introduction}
\label{sec:intro}

\begin{figure*}[ht]
    \centering
    \includegraphics[width=\linewidth]{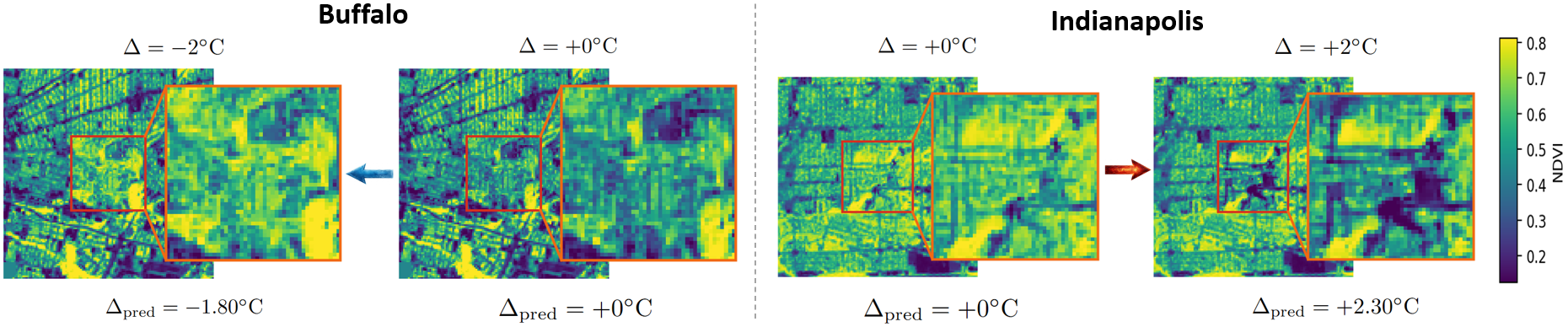}
    \caption{
    Our conflated framework combines a predictive forward model with a diffusion-based generative inverse model to produce diverse, physically plausible image-based vegetation patterns conditioned on specific temperature goals (left pair temperature decrease and right pair temperature increase). Higher NDVI values (closer to 1) indicate denser vegetation, while lower NDVI values indicate sparser vegetation.
    }
    \label{fig:teaser}
\end{figure*}

Urban areas are increasingly affected by climate variability and thermal extremes driven by rapid urbanization and global climate change~\cite{Masson2020,Peng2012}. Vegetation plays a central role in regulating urban microclimates through shading, evapotranspiration, and surface energy exchange~\cite{Aram2019}. Earth observation satellites (e.g., Landsat) enable large-scale urban measurements~\cite{Loveland2012}. Land surface temperature (LST) is estimated using a single-channel LST retrieval grounded in thermal radiative transfer theory. LST retrieval is a function of surface properties and emitted thermal radiance~\cite{Chander2009}. The Normalized Difference Vegetation Index (NDVI), an index derived from red and near-infrared reflectance to quantify vegetation presence and density, is incorporated to represent surface characteristics when converting brightness temperature to LST. Hence, the approach is a forward model, as it relies on the causal relationship between surface properties and the emitted radiance to retrieve the temperature~\cite{Sobrino2004LST}. %

A critical challenge for urban planning is determining vegetation configurations that induce a desired regional temperature shift through inverse LST modeling. Unlike forward temperature prediction, this vegetation-based temperature modulation problem remains largely unexplored. The task is inherently underdetermined, as multiple spatial vegetation arrangements can produce similar aggregated temperature responses within a neighborhood. To illustrate the ambiguity of the task, we partitioned urban regions into bins based on building height and LST profiles; even within self-similar bins, NDVI configurations varied by 24\% (e.g., a standard deviation of 0.16 over a range of 0.67; see Supp. 1). Deterministic regression models and conventional neural networks are poorly suited to represent this ambiguity, as they tend to converge to averaged spatial patterns with limited variability. The challenge is compounded by data scarcity, since we do not observe the same urban area under multiple vegetation scenarios and temperature-modification goals. 

Our key idea is to develop a deep learning model that is explicitly encouraged to produce diversity (e.g., spatially distinct vegetation patterns) as well as specificity (e.g., that all generated results satisfy a specified target condition such as a desired temperature change), even in data-scarce settings (Figure~\ref{fig:teaser}). The conflated result is the ability to generate diverse NDVI patterns for a single urban area that all meet a desired temperature objective—even when such combinations are absent from the training data.

\begin{figure*}[ht]
    \centering
    \includegraphics[width=0.8\linewidth]{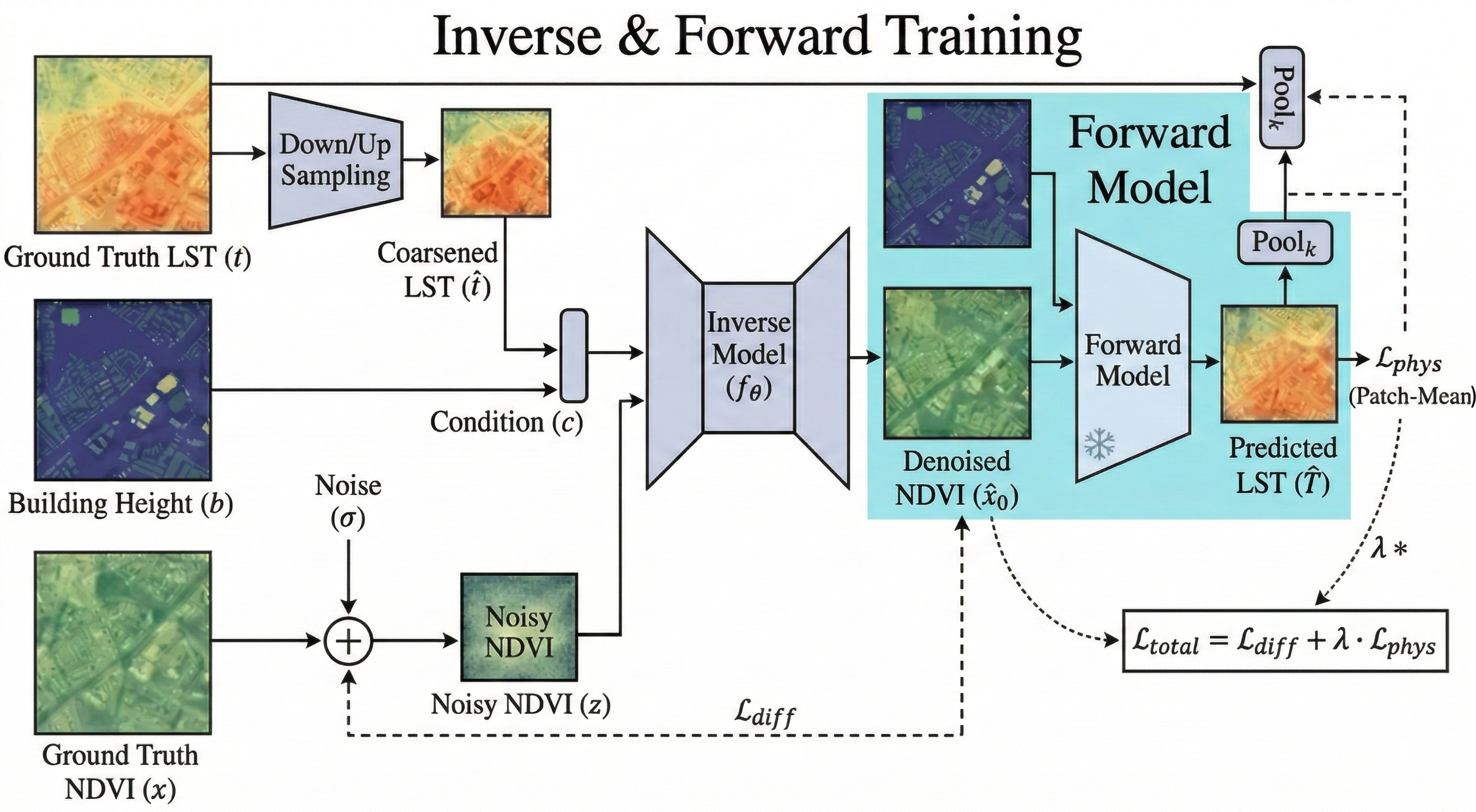}
    \caption{
    We show our system pipeline, including inverse and forward model, and the training and inference processes.
    }
    \label{fig:pipeline}
\end{figure*}

Our framework includes a learning process for a predictive forward model, an inference process for a generative inverse model, and a training procedure that integrates both (Figure~\ref{fig:pipeline}). We partition satellite imagery (Landsat 8) of multiple urban areas into 3.84 by 3.84 km tiles, each containing NDVI, LST, and building-height (BH) data. A U-Net–based~\cite{ronneberger2015unet} forward model is trained to predict LST from NDVI and BH inputs. A diffusion-based inverse model~\cite{karras2022edm} is trained to generate NDVI conditioned on BH and supervised by LST at an aggregated regional scale. During inverse model training, discrepancies between regional mean temperature predicted by the forward model and corresponding ground-truth temperature are penalized. At inference, users specify desired regional temperature changes as coarse conditioning inputs. By enforcing temperature at this aggregated level, the model retains flexibility to generate diverse spatial configurations within each region while maintaining control over regional temperature outcomes.

We applied our method to 20 US cities (listed in Supp. S2), spanning 41715 sq. kms. Results include training analysis, comparisons, ablation studies, and multiple inference results showing both diversity and specificity. Our approach increases diversity by 3.4 times and reduces specificity error by 37\% over baseline methods.

Our contributions include:
\begin{itemize}
\item{We introduce a combined forward and inverse model that achieves the conflated goal of both diverse output and specific temperature-inducing outputs.}
\item{We formulate vegetation-driven temperature modulation as a generative inverse modeling problem using NDVI, BH, and LST maps.}
\item{We show that directly conditioning diffusion models on fine-resolution temperature maps leads to limited spatial diversity and weaker regional temperature control.}
\end{itemize}

\section{Related Work}
\label{sec:relatedwork}

\paragraph{Urban Sensing.}
Urban thermal remote sensing has long been used to characterize urban surface temperature patterns using satellite-derived land surface temperature (LST) products \cite{marzban2018influence, logan2020night, duncan2019turning}. The presence of vegetation, commonly quantified through NDVI \cite{Huete1997, Anbazhagan2016}, is consistently associated with lower LST due to radiative shading and evapotranspiration effects \cite{Litvak2017}. However, the NDVI–LST relationship is nuanced; it varies with land cover composition \cite{Xue2021}, urban morphology \cite{Lindberg2011}, seasonality, and spatial aggregation scale \cite{Touhami2022}. Moreover, LST is a radiometric surface measure rather than near-surface air temperature, and thus serves as a proxy for surface energy balance rather than direct human thermal exposure \cite{Voogt2003}.

These observations highlight an important ambiguity: similar aggregated thermal statistics can arise from multiple fine-scale vegetation configurations. While NDVI and LST are strongly linked, LST alone does not uniquely determine the spatial arrangement of vegetation that produced it, especially in heterogeneous urban environments \cite{Muse2024}.

\paragraph{Predictive Greening-Based Heat Mitigation.}
A large body of work models LST as a prediction target using vegetation indices, albedo, built-up indices, and related surface descriptors \cite{Varamesh2022,Suthar2024}. Machine learning approaches, including tree-based ensembles, often achieve strong predictive performance, with NDVI emerging as a dominant explanatory variable \cite{Krishnaswamy2004}. Urban morphology, particularly three-dimensional structure such as building height, further modulates surface thermal patterns and is increasingly incorporated using large-scale building-height datasets \cite{Che2025,Stewart2012}. These studies adopt a forward perspective: given land cover and morphology, predict temperature.

Complementary work addresses planning by optimizing the spatial allocation of greening to reduce heat exposure or LST extremes \cite{Balany2020,Imran2021}. Such approaches demonstrate that targeted vegetation increases can outperform uniform treatments at aggregated scales \cite{Massaro2023}. But, most optimization and planning frameworks operate on coarse decision variables and return a single deterministic solution. They do not explicitly model the one-to-many nature of feasible fine-scale vegetation layouts that satisfy the same regional thermal objective. This leaves a gap between forward temperature prediction and generative design: how to produce multiple plausible vegetation configurations that achieve a specific regional temperature shift under morphological constraints.

\paragraph{Diffusion Models for Controllable and Inverse Generation.}
Diffusion models were introduced as denoising probabilistic generative models~\cite{ho2022classifierfree} and later formalized within score-based stochastic differential equation frameworks~\cite{song2021scorebased}, providing stable training and flexible conditioning mechanisms. Recent refinements such as EDM~\cite{karras2022edm} clarify preconditioning and sampling design choices that improve efficiency and robustness~\cite{ho2020denoising}. In remote sensing, diffusion-based models have been applied to satellite image synthesis and conditional tasks such as inpainting and multi-spectral reconstruction. DiffusionSat demonstrates that diffusion priors can capture the statistics of multispectral Earth observation imagery~\cite{khanna2024diffusionsat}.

Data-driven architectures, such as the adaptive Fourier neural operators utilized in FourCastNet \cite{pathak2022fourcastnet}, have demonstrated the massive potential of machine learning in high-resolution, forward weather modeling. More recently, the field has rapidly shifted toward probabilistic generation using diffusion-based architectures. Models like GenCast \cite{price2023gencast} and deterministic guidance-based diffusion frameworks \cite{hua2024weather, yoon2024probabilistic} have established a new standard for ensemble forecasting by generating diverse, physically realistic atmospheric states rather than single deterministic trajectories. However, their capability as an \textit{inverse} tool for urban climate adaptation, specifically, sampling diverse spatial vegetation patterns conditioned on specific temperature changes, remains an open challenge.

Controllability in diffusion models is often achieved through guidance strategies that balance condition adherence and diversity~\cite{dhariwal2021diffusion}. Conditioning on coarse or low-frequency structure has been shown to reduce over-determinism in one-to-many mappings (e.g., ILVR~\cite{ezaki2021ilvr}). Diffusion can be a solver for inverse problems by combining a learned prior with a forward model and enforcing measurement consistency through gradient-based guidance or posterior sampling~\cite{chung2023diffusion}. Strict pixel-level constraints can over-constrain generation under model mismatch, motivating softer constraint formulations and localized editing strategies such as diffusion-based inpainting~\cite{lugmayr2022repaint}.

Building on these ideas, we treat vegetation-driven temperature modulation as a generative inverse problem and enforce thermal consistency at an aggregated regional scale, aiming to preserve spatial diversity while achieving specific temperature outcomes.

\section{Methodology}
\label{sec:method}

\subsection{Problem Setup}
We study inverse modeling of urban vegetation patterns represented by NDVI images.
Let $\mathbf{x}\in\mathbb{R}^{1\times H\times W}$ be the target NDVI image tile and let the conditioning be
$\mathbf{c}=[\mathbf{b},\tilde{\mathbf{t}}]\in\mathbb{R}^{2\times H\times W}$,
where $\mathbf{b}$ is the geolocated building height (BH) map and $\tilde{\mathbf{t}}$ is a \emph{coarsened} geolocated land surface temperature (LST) map (Sec.~\ref{sec:cond}). We use building height as a proxy for urban morphology, as it captures building presence and influences street orientation, shade patterns, and heat retention.

Our goal is to learn a conditional generative model $p_\theta(\mathbf{x}\mid \mathbf{c})$, parameterized by $\theta$, that can sample diverse NDVI maps consistent with coarse thermal structure.

\subsection{Learned Forward Model}
\label{sec:fwd}
To assist our inverse model training, we use a learned forward model $g_\phi$ that predicts temperature \emph{change} from NDVI and building height (BH):
\begin{equation}
\Delta \hat{\mathbf{T}} = g_\phi\big(\hat{\mathbf{x}}_0,\mathbf{b}\big),
\end{equation}
where $\hat{\mathbf{x}}_0$ is the denoised NDVI prediction from Eq.~\ref{eq:x0hat}, and $\mathbf{b}$ denotes the building height (BH) map.
Absolute temperature is recovered by adding a per-tile baseline temperature 
$T_{\mathrm{base}}$, computed as the mean LST of the conditioning tile:
\begin{equation}
\hat{\mathbf{T}} = T_{\mathrm{base}} + \Delta \hat{\mathbf{T}}.
\end{equation}

We implement $g_\phi$ as a U-Net using \texttt{segmentation\_models\_pytorch} with a ResNet-50 encoder and ImageNet-pretrained encoder weights. $g_\phi$ is trained separately and frozen during subsequent inverse model training.

\subsection{Inverse Modeling Training}

\subsubsection{Conditional Diffusion Model}
\label{sec:edm}
{We implement the inverse model following the Score-SDE architecture~\cite{song2021scorebased} with an NCSN++ backbone, while replacing the original score-matching loss and sampling procedure with the EDM formulation~\cite{karras2022edm}. We later incorporate the learned forward model into training. Full architecture and training details are provided in Supp. S3.

Given $(\mathbf{x},\mathbf{c})$ where $\mathbf{x}$ is the target spatial NDVI pattern and $\mathbf{c}$ is the spatial condition, we sample a noise level $\sigma$ from a log-normal distribution and corrupt only the target:
\begin{equation}
\mathbf{z}=\mathbf{x}+\sigma\boldsymbol{\epsilon},\quad \boldsymbol{\epsilon}\sim\mathcal{N}(\mathbf{0},\mathbf{I}).
\end{equation}
The network $f_\theta$ concatenates the scaled noisy target and clean conditioning and predicts a denoising residual:
\begin{equation}
\hat{\mathbf{x}}_0 = c_{\mathrm{skip}}(\sigma)\mathbf{z} + c_{\mathrm{out}}(\sigma)\, f_\theta\big(c_{\mathrm{in}}(\sigma)\mathbf{z}\;\Vert\;\mathbf{c},\,\sigma\big).
\label{eq:x0hat}
\end{equation}
where the preconditioning coefficients are
\begin{align}
c_{\mathrm{in}} &= (\sigma^2+\sigma_{\mathrm{data}}^2)^{-1/2}, \\
c_{\mathrm{skip}} &= \frac{\sigma_{\mathrm{data}}^2}{\sigma^2+\sigma_{\mathrm{data}}^2}, \\
c_{\mathrm{out}} &= \frac{\sigma\,\sigma_{\mathrm{data}}}{\sqrt{\sigma^2+\sigma_{\mathrm{data}}^2}}.
\end{align}

The inverse model loss function, so far, follows the EDM weighted denoising formulation:
\begin{equation}
\mathcal{L}_{\mathrm{diff}}=
\mathbb{E}\left[w(\sigma)\,\lVert \hat{\mathbf{x}}_0-\mathbf{x}\rVert_2^2\right],\quad
w(\sigma)=\frac{\sigma^2+\sigma_{\mathrm{data}}^2}{(\sigma\sigma_{\mathrm{data}})^2}.
\end{equation}

\subsubsection{Coarsened LST Conditioning}
\label{sec:cond}
We coarsen LST before using it for conditioning by downsampling and then upsampling. In our preliminary experiments, we observe that using fine-resolution LST as a conditioning map causes an inverse model to produce an almost deterministic output -- the model produces NDVI patterns closely aligned with fine LST gradients. A coarsening process gives the model the freedom to vary the output patterns. Our downsampling and upsampling produces a coarsened temperature:
\begin{equation}
\tilde{\mathbf{t}} = \mathrm{Up}\big(\mathrm{Down}(\mathbf{t};k);k\big),
\end{equation}
where $\mathrm{Down}(\cdot;k)$ reduces spatial resolution by a factor $k$ and $\mathrm{Up}(\cdot;k)$ resizes back to $H\times W$ using interpolation. This preserves coarse regional temperature structure while removing fine-scale cues.

\subsubsection{Coarse Patch-Mean Physics Loss}
\label{sec:phys}

Next, we also desire the mean temperature of a patch (predicted from NDVI using $g_\phi$) to be equal to the mean temperature of the ground truth temperature of the same patch region. For this, let $\mathrm{Pool}_k(\cdot)$ denote non-overlapping $k\times k$ average pooling (stride $k$), applied after cropping to the largest multiple of $k$:
\begin{equation}
\bar{\mathbf{T}} = \mathrm{Pool}_k(\mathbf{T}),\quad
\bar{\hat{\mathbf{T}}}=\mathrm{Pool}_k(\hat{\mathbf{T}}).
\end{equation}

The inverse model loss is then augmented with an $\ell_1$ physics penalty on the pooled residual using the equation:
\begin{equation}
\mathcal{L}_{\mathrm{phys}} = \left\lVert \bar{\hat{\mathbf{T}}}-\bar{\mathbf{T}}\right\rVert_1.
\label{eq:lphys}
\end{equation}
We use an $\ell_1$ penalty to provide stable gradients for small but systematic temperature errors and to reduce sensitivity to outliers as compared to the behavior of $\ell_2$.

\subsubsection{Total Loss Function and Lambda Schedule}
The full training loss function is
\begin{equation}
\mathcal{L} = \mathcal{L}_{\mathrm{diff}} + \lambda_{\mathrm{phys}}(s)\,\mathcal{L}_{\mathrm{phys}},
\end{equation}
where $\lambda_{\mathrm{phys}}(s)$ follows a warmup then linear ramp to a maximum value $\lambda_{\max}$ over training step $s$:
\begin{equation}
\lambda_{\mathrm{phys}}(s)=
\begin{cases}
0, & s < s_{\mathrm{warm}},\\
\lambda_{\max}\frac{s-s_{\mathrm{warm}}}{s_{\mathrm{ramp}}}, & s_{\mathrm{warm}}\le s < s_{\mathrm{warm}}+s_{\mathrm{ramp}},\\
\lambda_{\max}, & s \ge s_{\mathrm{warm}}+s_{\mathrm{ramp}}.
\end{cases}
\end{equation}

This schedule stabilizes training by allowing the diffusion model to first learn basic denoising behavior before enforcing temperature consistency. We choose $s_{\mathrm{warm}}$ and $s_{\mathrm{ramp}}$ to delay and then gradually introduce the physics-based constraint, avoiding abrupt optimization changes early in training.

\subsection{Inverse Model Inference}
\label{sec:sampling}
At inference time, we sample with an EDM solver using a Karras noise schedule~\cite{karras2022edm} $\{\sigma_i\}_{i=0}^{S}$ with $\sigma_S=0$.
We initialize the editable NDVI region with Gaussian noise at $\sigma_{\max}$ and keep the coarsened conditioning $\mathbf{c}$ fixed.

\paragraph{Inpainting constraint.}
We enforce that NDVI is modified only inside the editable region, which implicitly ensures the modification is compatible with the surroundings. Let $\mathbf{M}\in\{0,1\}^{1\times H\times W}$ be an edit mask where $\mathbf{M}=1$ indicates editable pixels.
At each step with noise level $\sigma$, we project the current sample to preserve known pixels by replacing $(1-\mathbf{M})$ with a noisy reference:
\begin{equation}
\mathbf{x}\leftarrow
\mathbf{M}\odot \mathbf{x} + (1-\mathbf{M})\odot (\mathbf{x}_{\mathrm{ref}} + \sigma\boldsymbol{\eta}),
\quad \boldsymbol{\eta}\sim\mathcal{N}(\mathbf{0},\mathbf{I}).
\label{eq:inpaint}
\end{equation}

\paragraph{EDM update.}
Given $\mathbf{x}_i$ at noise level $\sigma_i$, we compute $\hat{\mathbf{x}}_0$ via Eq.~\ref{eq:x0hat} and perform an Euler update in data space:
\begin{equation}
\mathbf{d}_i=\frac{\mathbf{x}_i-\hat{\mathbf{x}}_0}{\sigma_i},\quad
\mathbf{x}_{i+1}=\mathbf{x}_i+(\sigma_{i+1}-\sigma_i)\mathbf{d}_i,
\end{equation}
followed by the projection in Eq.~\ref{eq:inpaint}.

At inference time, temperature modification is specified by altering the LST condition within the editable region.
The diffusion model then synthesizes NDVI only inside this region via masked inpainting, while preserving the surrounding context.

\section{Experiments}
\label{sec:experiments}

\subsection{Experimental Setup}

\paragraph{Dataset and Temporal Split.}
We retrieve Landsat 8 Level-2 scenes for 20 U.S. cities using Google Earth Engine.
Land Surface Temperature (LST) is generated from Thermal Band 10 (TIR), 
and NDVI is computed from Landsat 8 surface reflectance bands 4 (red) and 5 (near-infrared, NIR), all extracted from the same acquisition scenes to ensure temporal alignment. Building height information is derived from the US Building Height dataset \cite{Che2025}.

Each city is determined using its administrative boundary (derived from TIGER database~\cite{census_tiger_2024}). From each cropped scene, we extract non-overlapping $128 \times 128$ tiles at 30m spatial resolution (i.e., each pixel represents a $30\text{m} \times 30\text{m}$ area). After filtering, the dataset contains: 2829 tiles for training, 701 for testing.

Training and test splits share the same set of cities but use different acquisition dates, enabling evaluation of temporal generalization within known urban morphologies. Training is performed in PyTorch on a single NVIDIA RTX 6000 Ada GPU and typically takes about 4 hours for a single model configuration.

\paragraph{Diffusion Training Details.}
We train the EDM model for 24000 iterations using Adam with a learning rate of $2 \times 10^{-4}$.
The LST coarsening factor is $k = 32$.
The pooling size for the physics loss is $k_{\text{pool}} = 32$.
The maximum physics weight is $\lambda_{\max} = 16$.

\paragraph{Inference Protocol.}
At inference time, we modify vegetation only within a fixed 
$32 \times 32$ pixel region of interest (ROI) inside each 
$128 \times 128$ tile.
Given the 30m spatial resolution of Landsat 8, this corresponds to an area of approximately $0.96 \text{ km} \times 0.96 \text{ km}$, 
representing neighborhood-scale interventions rather than trivial 
global modifications or unrealistically small micro-scale changes.

We set the LST coarsening factor to $k=32$ and the physics pooling size to $k_{\text{pool}}=32$ so that both conditioning and physics supervision operate at the same spatial scale as the $32\times32$ intervention ROI. This design encourages the model to match neighborhood-scale temperature behavior rather than overfitting to fine-scale LST details outside the intended intervention scale.

We change the LST condition only inside the ROI by
\[
\Delta_{\text{cond}} = w \cdot \Delta_{\text{target}},
\]
while keeping the surrounding context unchanged.
The diffusion model then generates NDVI within the ROI using the sampling mechanism described in Sec.~\ref{sec:sampling}. Similar to guidance scaling in conditional diffusion models~\cite{ho2022classifierfree}, we modulate the magnitude of the induced temperature change by altering $w \in \{1,2,3,5,8\}$.

\subsection{Evaluation Metrics}

We define the following four complementary metrics.
\vspace{-0.15in}

\begin{figure*}[t]
    \centering
    \includegraphics[width=\textwidth]{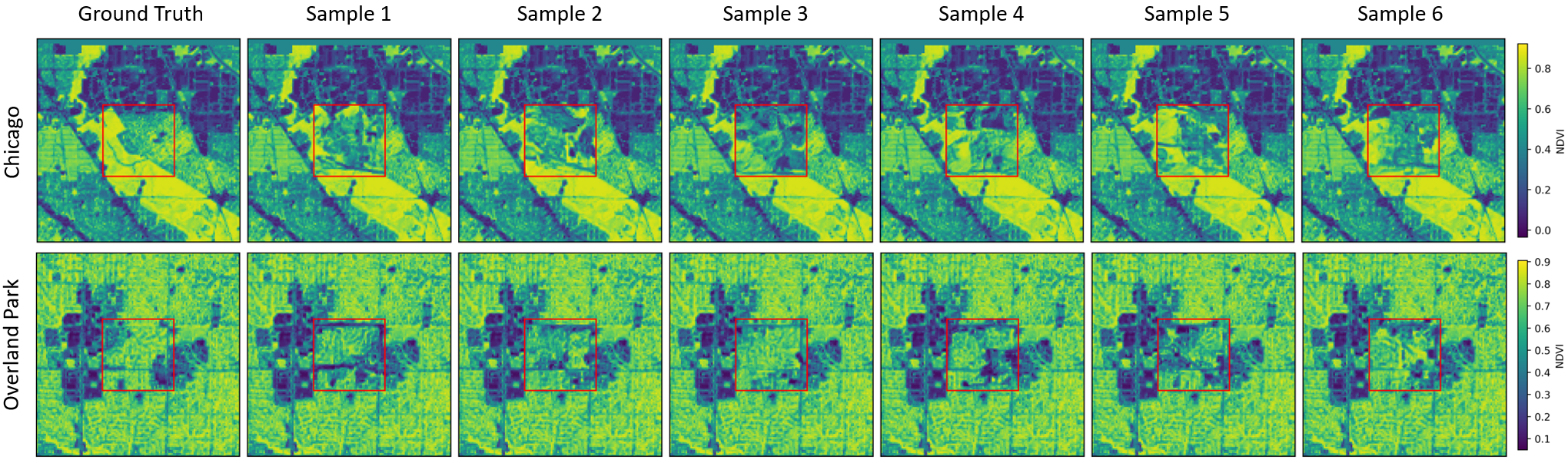}
    \caption{
       \textbf{Diversity.} Diverse NDVI generation under identical conditioning ($\Delta$ = 0) for two representative cities (Chicago and Overland Park); images demonstrate consistent diversity across distinct urban morphologies while preserving context outside the ROI. Higher NDVI values (closer to 1) indicate denser vegetation, while lower NDVI values indicate sparser vegetation.
    }
    \label{fig:diversity}
\end{figure*}

\begin{figure*}[t]
    \centering
    \includegraphics[width=1.0\textwidth]{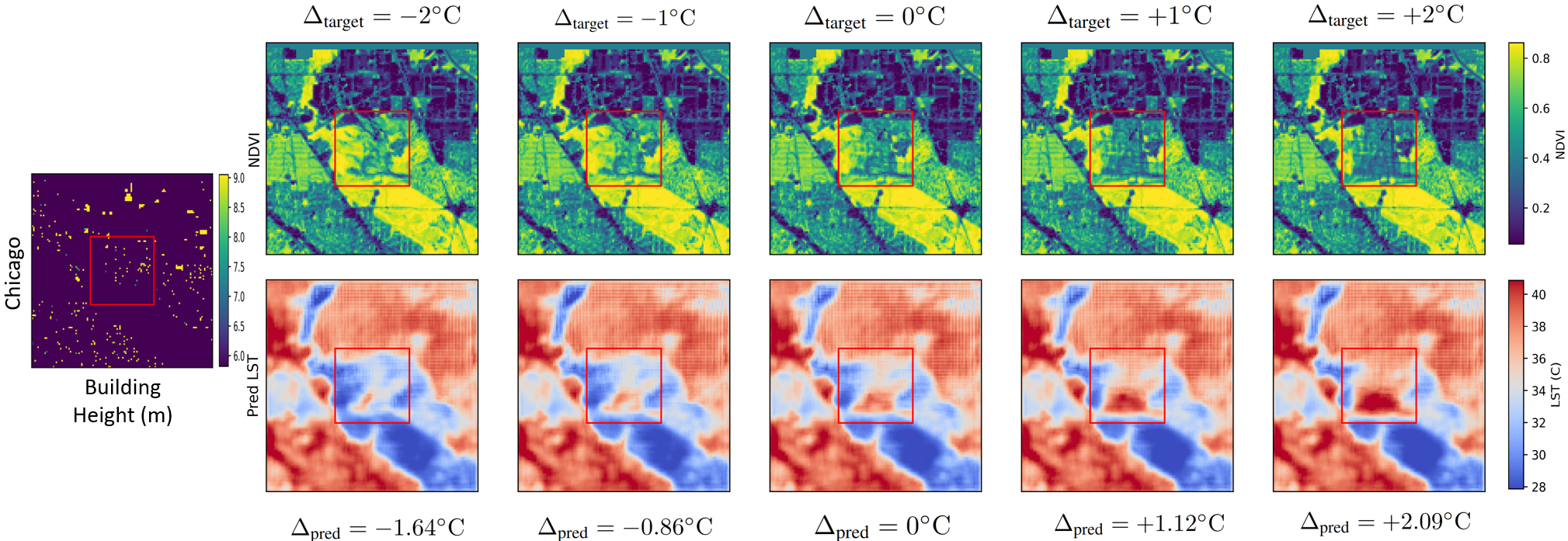}
    \caption{
   \textbf{Chicago Specificity.} Images show the ability to alter regional temperature to specific changes in a Chicago tile.
    Top: generated NDVI for 
    $\Delta_{\text{target}} \in \{-2,-1,0,+1,+2\}^\circ$C.
    Bottom: forward-model predicted LST.
    Numbers indicate ROI mean $\Delta_{\text{pred}}$.
    }
    \label{fig:delta_sweep_chicago}
\end{figure*}

\begin{figure*}[t]
    \centering
    \includegraphics[width=1.0\textwidth]{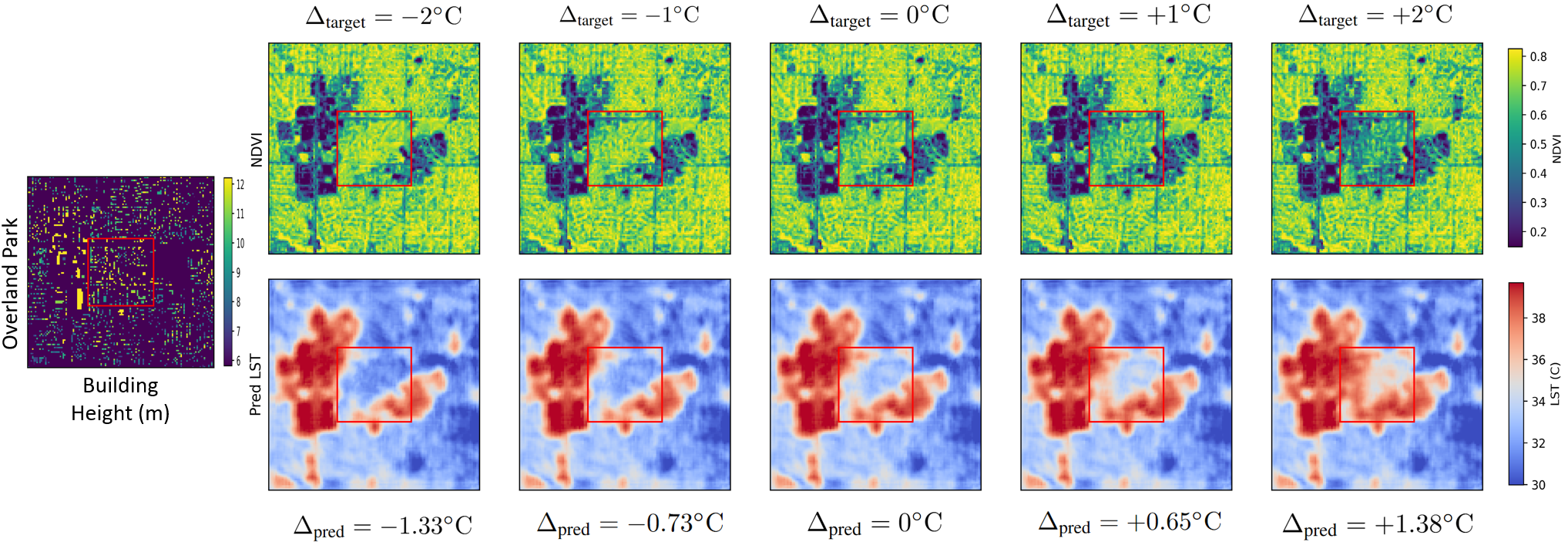}
    \caption{
    \textbf{Overland Park Specificity.} Images show the ability to alter regional temperature to specific changes in an Overland Park tile.
    Top: generated NDVI for 
    $\Delta_{\text{target}} \in \{-2,-1,0,+1,+2\}^\circ$C.
    Bottom: forward-model predicted LST.
    Numbers indicate ROI mean $\Delta_{\text{pred}}$.
    }
    \label{fig:delta_sweep_overlandpark}
\end{figure*}

\paragraph{a. Temperature Control Error (CtrlErr).}

Control error measures whether the requested temperature change is achieved 
within the region of interest (ROI).
For a target temperature change $\Delta_{\text{target}}$, we first compute the predicted 
temperature difference relative to the generated $\Delta=0$ baseline:

\[
\Delta_{\text{pred}} =
\text{mean}_{\text{ROI}}\!\left(
T_{\text{pred}}(\Delta) - T_{\text{pred}}(0)
\right).
\]

The control error is then defined as

\[
\text{CtrlErr} =
|\Delta_{\text{pred}} - \Delta_{\text{target}}|.
\]

This metric evaluates the accuracy of achieving a temperature change (using a novel NDVI pattern). Since it is computed relative to the predicted baseline (i.e., zero change), it does not depend on ground-truth temperature values and isolates the model's ability to produce the desired change.

\paragraph{b. Baseline Consistency Error (BaseErr).}

Baseline consistency evaluates whether generated vegetation layouts produce temperatures that are similar to observed temperature distributions.
For samples generated with $\Delta=0$, we compare the predicted mean ROI temperature to the ground-truth mean ROI temperature:
\[
\text{BaseErr}_{\text{ROI}} =
\left|
\text{mean}_{\text{ROI}}\!\left(T_{\text{pred}}(0)\right)
-
\text{mean}_{\text{ROI}}\!\left(T_{\text{gt}}\right)
\right|.
\]

This metric measures how realistic the generated baseline vegetation is 
with respect to actual temperature observations. Unlike CtrlErr, it depends 
on ground-truth temperature and reflects absolute regional temperature correctness.

\paragraph{c. Surrogate Calibration Error (SurrCalErr).}
\label{sec:surr_cal_err}

Because temperature predictions rely on a learned forward model $g_\phi$, 
there exists an intrinsic calibration error independent of the diffusion model.
We therefore report the surrogate calibration error, defined as

\[
\hat{T}_{\text{ROI}}^{\text{gt}} =
\text{mean}_{\text{ROI}}\!\left(
T_{\text{base}} + g_\phi(\mathbf{x}_{\text{gt}}, \mathbf{b})
\right).
\]

\[
\text{SurrCalErr}_{\text{ROI}} =
\left|
\hat{T}_{\text{ROI}}^{\text{gt}}
-
\text{mean}_{\text{ROI}}\!\left(T_{\text{gt}}\right)
\right|.
\]

This quantity measures the forward model's error when evaluated on Landsat-derived NDVI. It defines a lower bound on achievable baseline 
consistency.

\paragraph{d. Diversity Metric.}
For each condition, we generate $N$ samples and compute within the ROI,
\[
\text{Diversity} =
\mathbb{E}_{i\neq j}[1 - \text{SSIM}(x_i, x_j)]
\]

\setlength{\tabcolsep}{3.5pt}
\begin{table}[t]
\centering
\caption{
\textbf{Training.} Results at ROI scale ($k=32$) and for various values for $\lambda$.
For each configuration, we report the best temperature control error 
(CtrlErr$_{\text{avg}}$) over $w \in \{1,2,3,5,8\}$. 
BaseErr$_{\text{ROI}}$ ($\Delta$ = 0) denotes the absolute difference between 
the predicted and ground-truth mean ROI temperature for samples 
generated with $\Delta$ = 0. 
}
\begin{tabular}{lccccccc}
\toprule
$\lambda$ & Best $w$ 
& CtrlErr$_{\text{avg}}$ 
& \begin{tabular}[c]{@{}c@{}}BaseErr$_{\text{ROI}}$\\$(\Delta = 0)$\end{tabular}
& Diversity \\
\midrule

1 & 5 
& $0.772 \pm 0.350$ 
& 1.49 
& $0.707 \pm 0.228$ \\

2 & 3 
& $0.715 \pm 0.301$ 
& 1.51 
& $0.830 \pm 0.118$ \\

4 & 3 
& $0.637 \pm 0.230$ 
& 1.53 
& $0.879 \pm 0.070$ \\

6 & 3 
& $0.609 \pm 0.251$ 
& 1.55 
& $0.888 \pm 0.064$ \\

8 & 3 
& $0.573 \pm 0.257$ 
& 1.59 
& $0.889 \pm 0.062$ \\

16 & 3 
& $0.559 \pm 0.310$
& 1.63
& $ 0.905 \pm 0.052 $ \\

32 & 3 
& $0.590 \pm 0.344$ 
& 1.71
& $ 0.905 \pm 0.053 $ \\

64 & 3 
& $0.633 \pm 0.282$ 
& 1.79
& $ 0.902 \pm 0.060 $ \\

\bottomrule
\end{tabular}
\label{tab:lambda_sweep}
\end{table}

Metrics are averaged over $K$ stochastic samples per condition, and we report mean and standard deviation.

\begin{table*}[t]
\centering
\caption{
\textbf{Comparisons.} We compare several configurations, each at the best $w$ and $\lambda$ value (if applicable). As observed, our method (last row) produces the lowest $CtrlErr$ and highest $Diversity$ values.
}
\begin{tabular}{lccccccc}
\toprule
Model & LST Coarse & Physics & Best $w$ 
& CtrlErr$_{\text{avg}}$ 
& BaseErr$_{\text{ROI}}$ ($\Delta$ = 0)
& Diversity \\
\midrule

U-Net
& \xmark & \xmark & 2 
& $0.211 \pm 1.097$
& 3.267 %
& - \\

Fine LST
& \xmark & \xmark & 3 
& $0.746 \pm 0.341$ 
& 1.50 
& $0.265 \pm 0.183$ \\

Coarse LST only 
& \cmark & \xmark & 8 
& $0.882 \pm 0.331$ 
& 1.57 
& $0.341 \pm 0.240$ \\

Coarse + Physics
& \cmark & \cmark & 3 
& $0.559 \pm 0.310$ 
& 1.633
& $ 0.905 \pm 0.052 $ \\

\bottomrule
\end{tabular}
\label{tab:compare}
\end{table*}

\begin{figure}[t]
    \centering
    \includegraphics[width=0.9\linewidth]{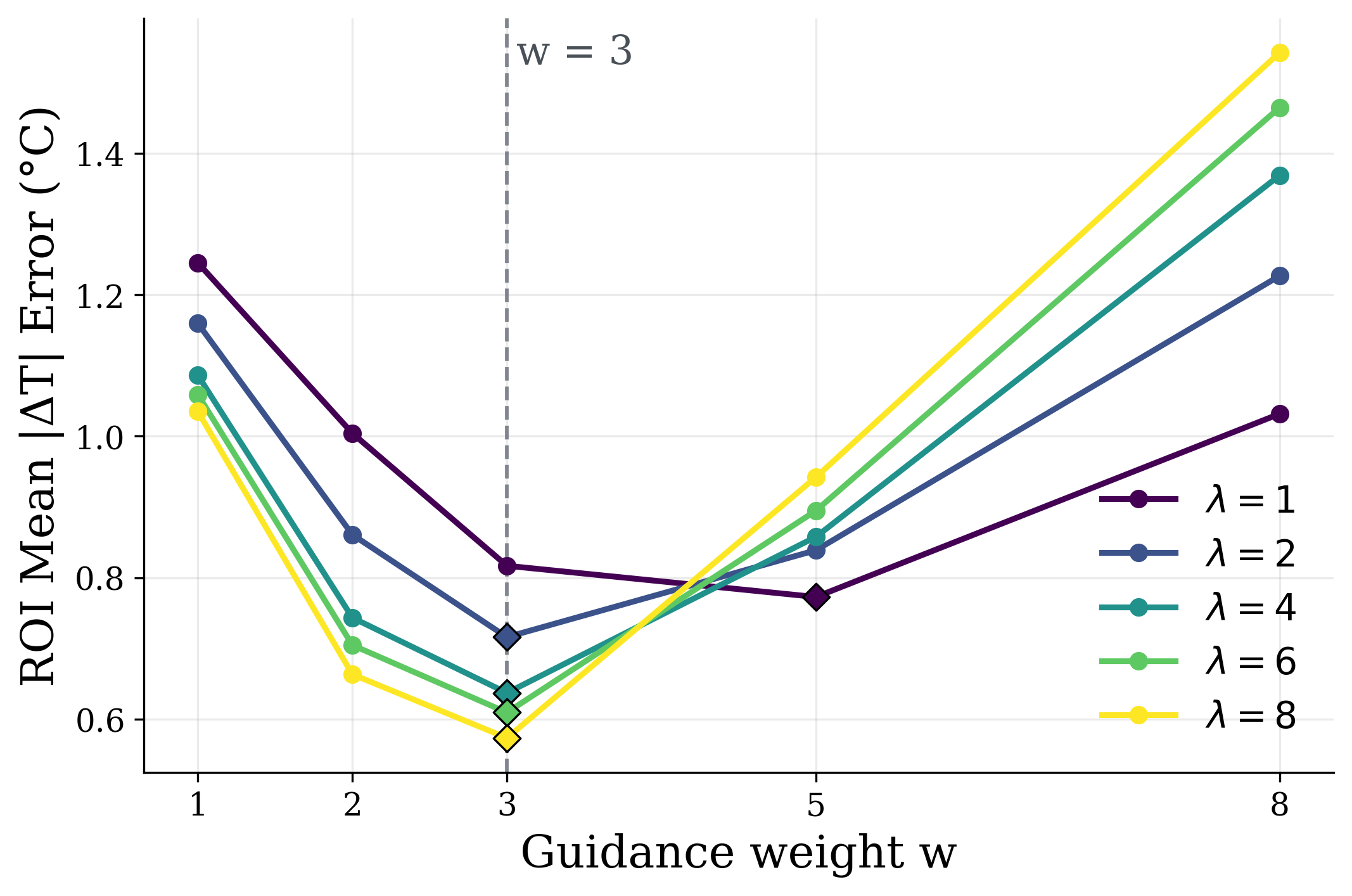}
    \caption{
    \textbf{Gain control.} Controllability as a function of $w$ gain.
    Y-axis shows ROI Mean $\Delta T$ Absolute Error (°C).
    Lower is better.
    }
    \label{fig:w_sweep}
\end{figure}

\begin{figure}[t]
    \centering
    \includegraphics[width=0.9\linewidth]{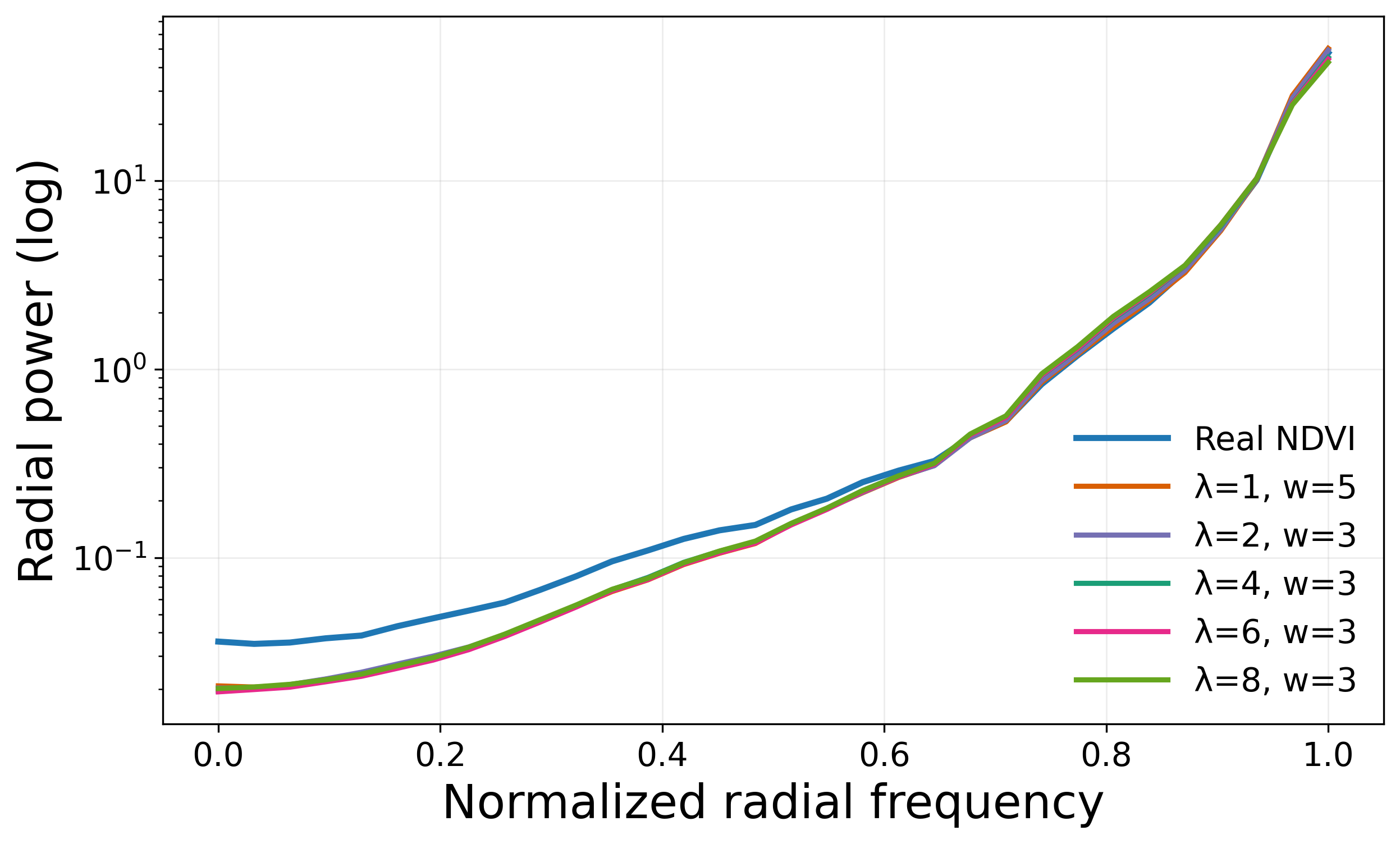}
    \caption{\textbf{Realism Proxy.} As a proxy for realism, we show radially averaged 2D Fourier power spectra of real and generated NDVI ROI patches. Power is averaged over frequency rings and plotted against normalized radial frequency (low: coarse structure; high: fine texture). Close alignment indicates similar spatial frequency statistics.}
    \label{fig:realism}

\end{figure}

\subsection{Training and Comparison Performance}

\textbf{Training.} We trained our inverse model in several configurations. Table~\ref{tab:lambda_sweep} reports the performance of our method using different $\lambda_{max}$ values. For each configuration, we choose the best performance $w$ gain value. The smallest CtrlErr occurs at $\lambda_{max}=16$. The BaseErr gradually increases with $\lambda_{max}$ as does diversity. Moderate values of $\lambda_{max}$ achieve the best trade-off between controllability and diversity, whereas excessively large values reduce variability. A good compromise performance is near $\lambda_{max}=16$ which we use for later reported results.

For the evaluation tiles used for reporting CtrlErr and BaseErr, the forward model error SurrCalErr$_{\text{ROI}}$ is 1.54$^\circ$C. We compute this calibration error on the same subset of tiles to provide a direct and fair lower bound for BaseErr$_{\text{ROI}}$, since both metrics are evaluated on identical conditions.

\textbf{Comparisons.} We also compare our method to several alternative approaches. In Table~\ref{tab:compare}, we show the performance of an end-to-end trained inverse U-Net with the same \texttt{segmentation\_models\_pytorch} ResNet-50 architecture used for the forward predictor, as well as EDM-based variants that share the same NCSN++ backbone and training setup described in Sec.~\ref{sec:edm}. These include an EDM model with typical fine-resolution LST conditioning and no physics loss, an EDM model with coarsened LST conditioning and no physics loss, and our best performing model. U-Net does not produce diversity. Fine-resolution conditioning leads to limited controllability and reduced diversity. Coarsening alone increases variability but weakens temperature alignment. Physics loss helps achieve the lowest control error while preserving diversity. Thus as shown our model outperforms in both diversity and specificity (i.e., CtrlErr).

\textbf{Ablation.} An ablation study of our method is implicitly performed in Table~\ref{tab:compare} and with Figure~\ref{fig:w_sweep}. Table~\ref{tab:compare}  shows the relative effect of different method components. Figure~\ref{fig:w_sweep} shows the effect of altering the gain $w$. The model exhibits under-response at low or high $w$ and improved controllability at intermediate values, consistent with guidance-scale behavior in diffusion models.

\subsection{Inference Performance}

At inference time, we demonstrate qualitatively the conflated goals of diversity and specificity. Figure~\ref{fig:diversity} illustrates the diversity of NDVI samples generated for the same temperature condition in two of our cities (Chicago and Overland Park). The proposed regional formulation produces diverse vegetation layouts within the ROI, while preserving surrounding context via masked inpainting. %

Figure~\ref{fig:delta_sweep_chicago} and Figure~\ref{fig:delta_sweep_overlandpark} show examples of producing temperature change inducing vegetation patterns in Chicago and Overland Park, respectively. Each column has both the target and predicted temperature change, which match well. Thus, considering Figure~\ref{fig:diversity} as well as Figure~\ref{fig:delta_sweep_chicago} and Figure~\ref{fig:delta_sweep_overlandpark}, our model can be used to obtain diverse vegetation patterns that may lead to desired urban temperature changes.

Figure~\ref{fig:realism} presents a proxy for the NDVI pattern quality. We observe high similarity between the spatial frequency statistics of the generated vs actual 2D NDVI spatial patterns. This helps confirm the realism of the produced NDVI images, at least at the scale of provided imagery.

Finally, in Section~S1 we show various NDVI patterns and temperature changes over additional cities, as produced automatically by our method with minimal user intervention (i.e., the user only needs to specify the input region and the desired temperature change).

\section{Conclusion}
\label{sec:conclusion}

We present a conflated inverse modeling framework for controllable and diverse generation of vegetation configurations under regional temperature targets. By coupling a U-Net-based forward LST predictor with a diffusion-based inverse generator and enforcing supervision at an aggregated temperature scale, our approach addresses the underdetermined nature of vegetation-driven temperature modulation. The framework overcomes limitations of deterministic regression by producing multiple spatially distinct NDVI patterns that satisfy the same thermal target even in data-scarce settings. Experiments across 20 U.S. cities show improved diversity while maintaining stronger temperature specificity than baseline approaches, demonstrating our approach as a scalable tool for urban climate adaptation and targeted vegetation design.

\noindent\textbf{Limitations and Future Work.}
In real-world settings, vegetation placement is constrained by existing infrastructure and architectural layouts. While our current framework considers only building constraints, future work could incorporate spatial masks and rule-based constraints into the diffusion process. Extending the analysis to additional satellite platforms would also strengthen generality. Sentinel-2 provides higher spatial resolution but lacks the thermal band required for LST estimation, whereas MODIS provides thermal data at a much coarser resolution. Given these trade-offs, Landsat provides a suitable balance for this study, though cross-sensor evaluation remains an important direction for future work.

\textbf{Acknowledgements.} This work was partially funded by NSF Award 2411273 and NSF Award 2107096.

{
    \small
    \bibliographystyle{ieeenat_fullname}
    \bibliography{main}
}

\clearpage
\setcounter{page}{1}
\maketitlesupplementary

\subsection*{S1. Additional Dataset-Level Analysis}

We provide an additional dataset-level analysis to illustrate the one-to-many nature of the inverse problem. Specifically, we measure tile-level NDVI variability after grouping tiles into similar building-height and mean-LST bins.

\begin{figure}[h]
    \centering 
    \includegraphics[width=1.0\columnwidth]{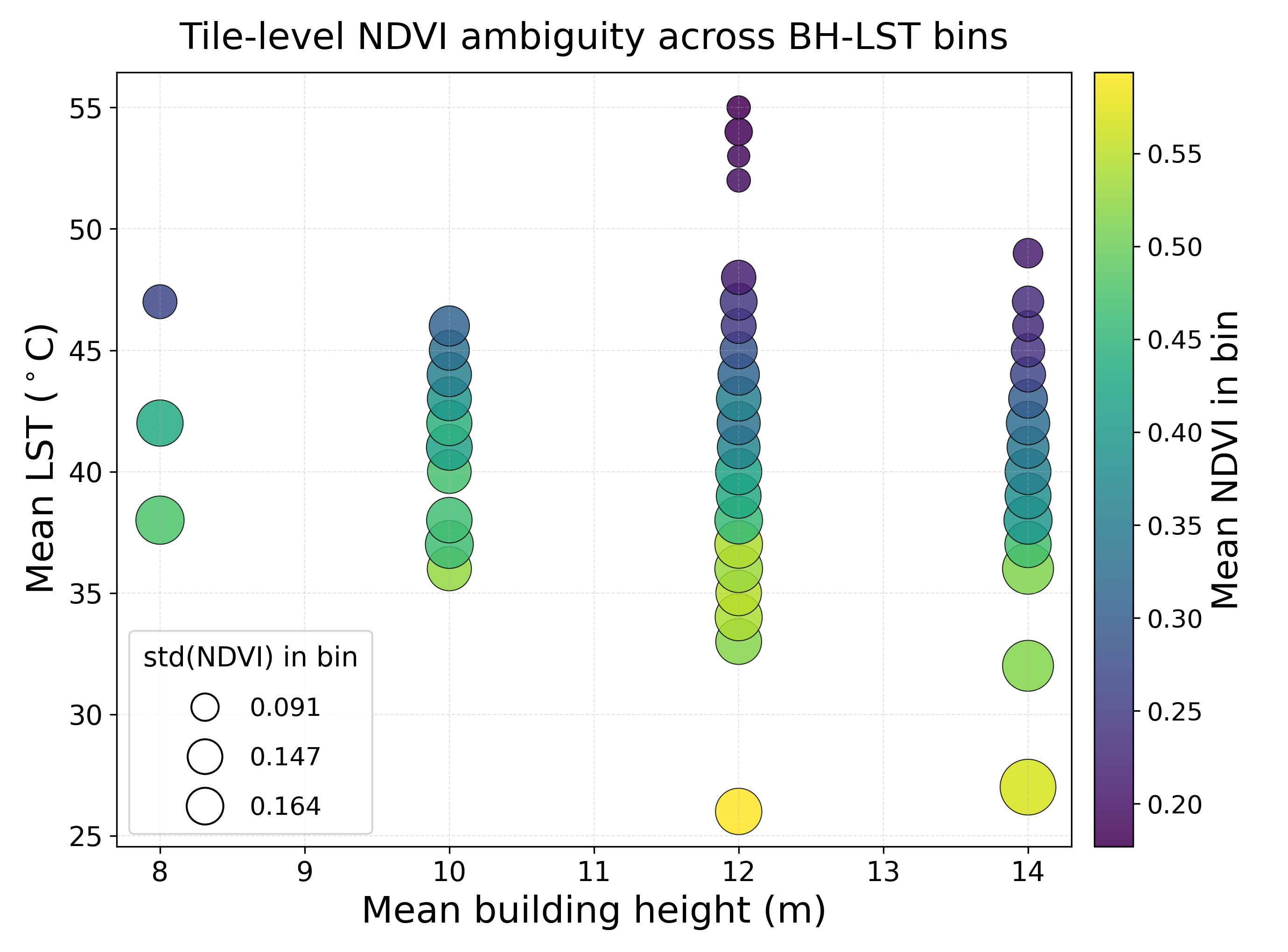}
    \caption{
    \textbf{NDVI Diversity.} We show tile-level NDVI variability observed in our dataset after binning tiles into similar building height and mean LST bins. Marker size corresponds to the standard deviation of NDVI within each bin (minimum count = 30). Standard deviations range from 0.09 to 0.16, indicating substantial within-bin variability.}
    \label{fig:ambiguity}
\end{figure}

\subsection*{S2. Additional Qualitative Results Across Cities}

We provide additional qualitative results across multiple cities in our dataset, shown in Fig.~\ref{fig:supp_city_rows}.
For each city, we show NDVI generation within the ROI under 
$\Delta \in \{-2,0,+2\}^\circ$C.
The reported values correspond to the ROI mean predicted temperature change 
$\Delta_{\text{pred}}$ obtained from the frozen forward model.

The 20 cities included in our dataset are:
Billings, MT; Bloomington, IN; Buffalo, NY; Chicago, IL; Denver, CO; Eugene, OR; 
Green Bay, WI; Indianapolis, IN; Jacksonville, FL; Los Angeles, CA; Norman, OK; 
Overland Park, KS; Phoenix, AZ; Pittsburgh, PA; Reno, NV; Salt Lake City, UT; 
San Jose, CA; Shreveport, LA; Sioux Falls, SD; and Syracuse, NY.

\begin{figure*}[t]
    \centering

    {\small \textbf{Buffalo}}\\
    \includegraphics[width=0.95\textwidth]{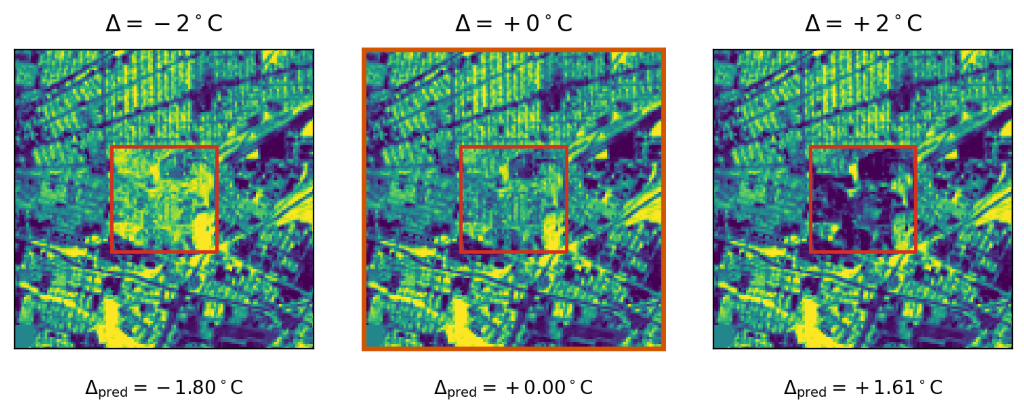}\vspace{4pt}

    {\small \textbf{Billings}}\\
    \includegraphics[width=0.95\textwidth]{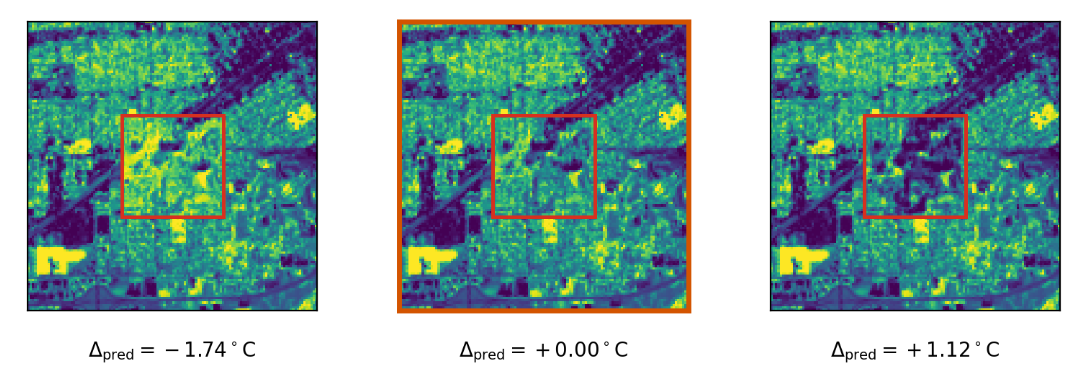}\vspace{4pt}

    {\small \textbf{Bloomington}}\\
    \includegraphics[width=0.95\textwidth]{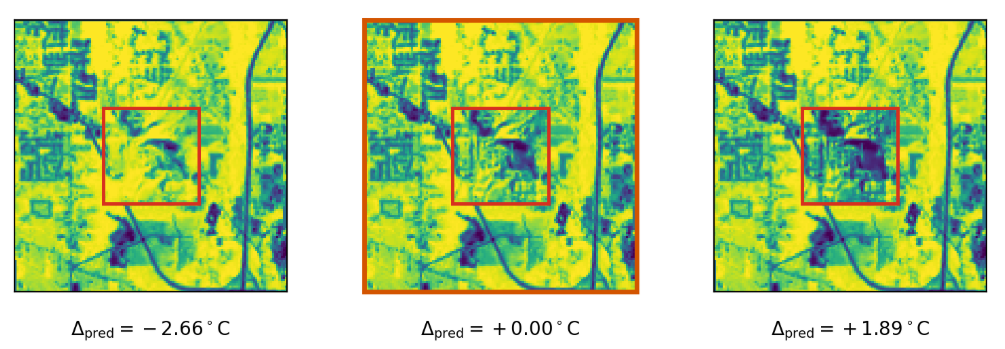}\vspace{4pt}

    \caption{
    Additional city-level qualitative results.
    Each row shows NDVI generation under $\Delta \in \{-2,0,+2\}^\circ$C with the ROI mean predicted change $\Delta_{\text{pred}}$.
    }
    \label{fig:supp_city_rows}
    \label{fig:supp_S1}
\end{figure*}

\begin{figure*}[t]
    \centering

    {\small \textbf{Denver}}\\
    \includegraphics[width=0.95\textwidth]{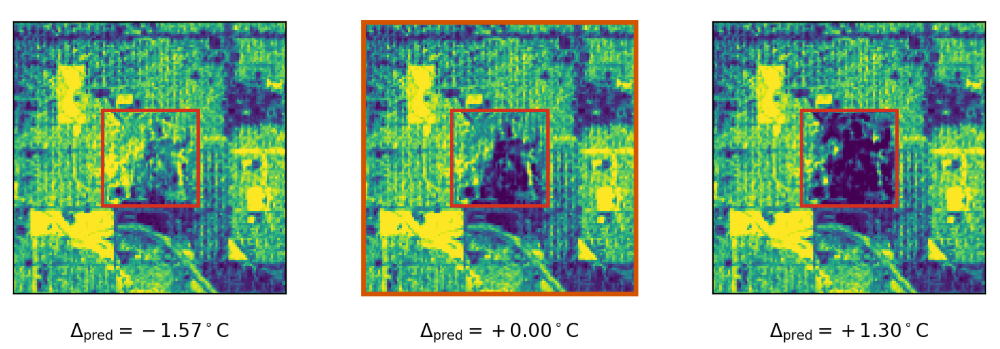}\vspace{4pt}

    {\small \textbf{Eugene}}\\
    \includegraphics[width=0.95\textwidth]{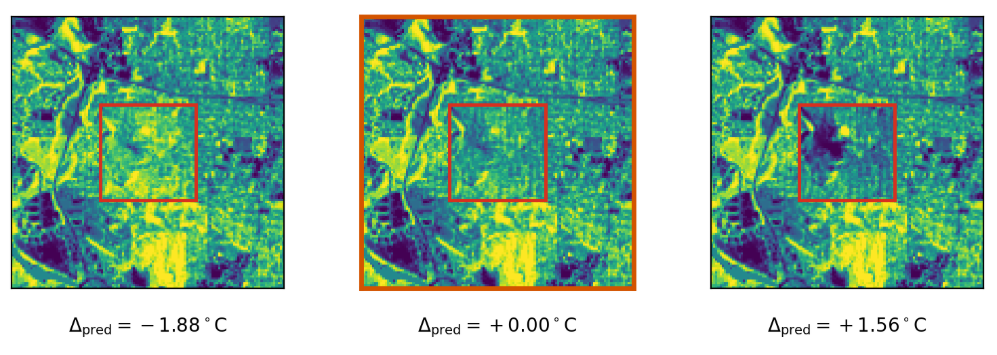}\vspace{4pt}

    {\small \textbf{Indianapolis}}\\
    \includegraphics[width=0.95\textwidth]{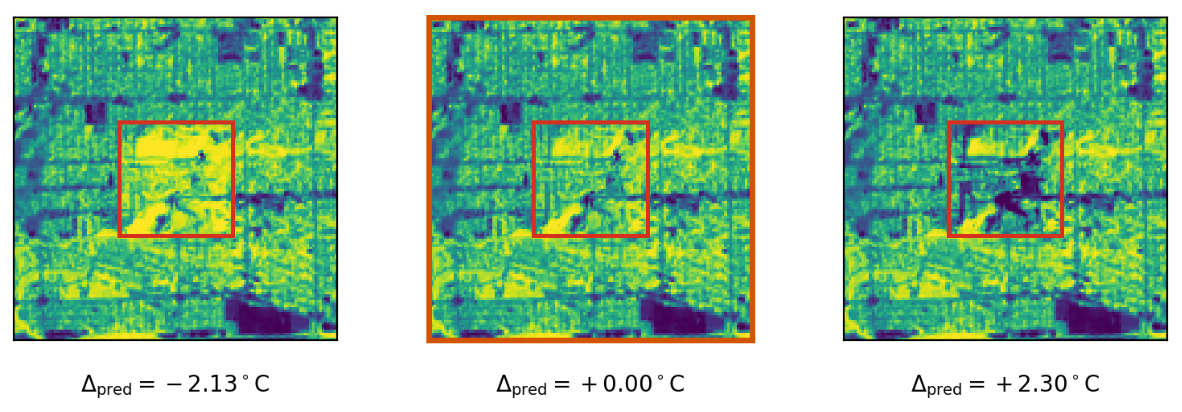}\vspace{4pt}

    \caption*{Figure~\ref{fig:supp_city_rows} (continued).}
\end{figure*}

\begin{figure*}[t]
    \centering

    {\small \textbf{Pittsburgh}}\\
    \includegraphics[width=0.95\textwidth]{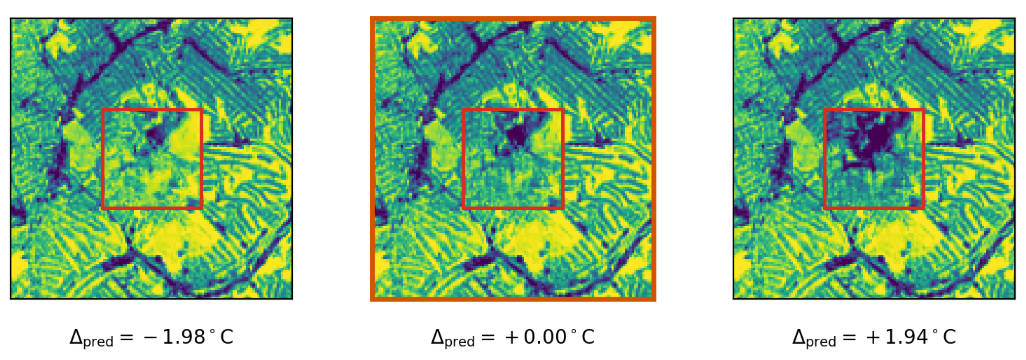}\vspace{4pt}

    {\small \textbf{Shreveport}}\\
    \includegraphics[width=0.95\textwidth]{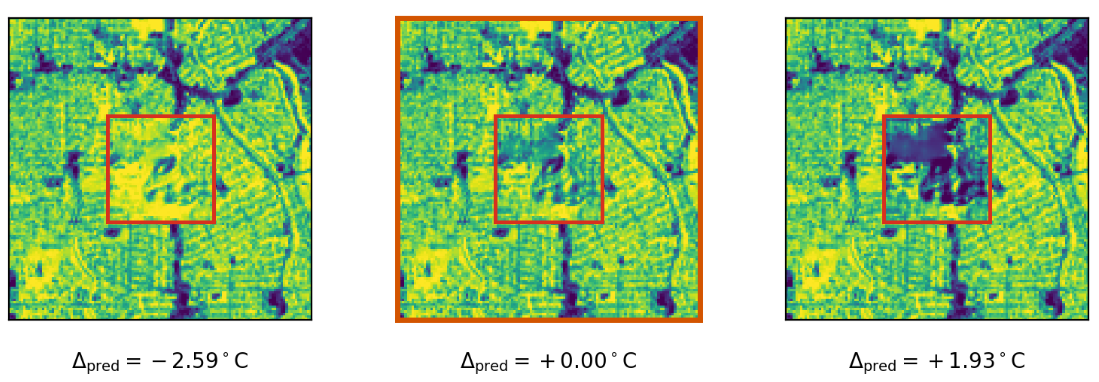}\vspace{4pt}

    {\small \textbf{Sioux Falls}}\\
    \includegraphics[width=0.95\textwidth]{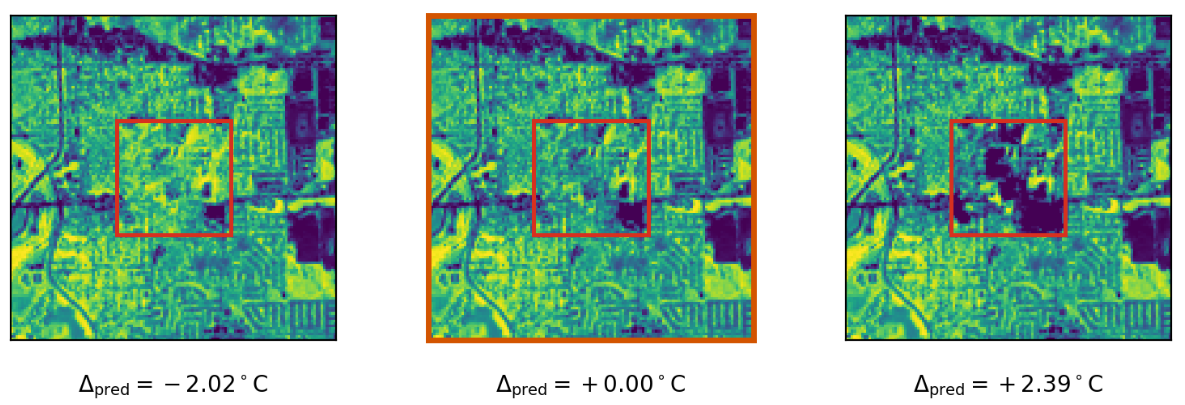}\vspace{4pt}

    \caption*{Figure~\ref{fig:supp_city_rows} (continued).}
\end{figure*}

\begin{figure*}[t]
    \centering

    {\small \textbf{Syracuse}}\\
    \includegraphics[width=0.95\textwidth]{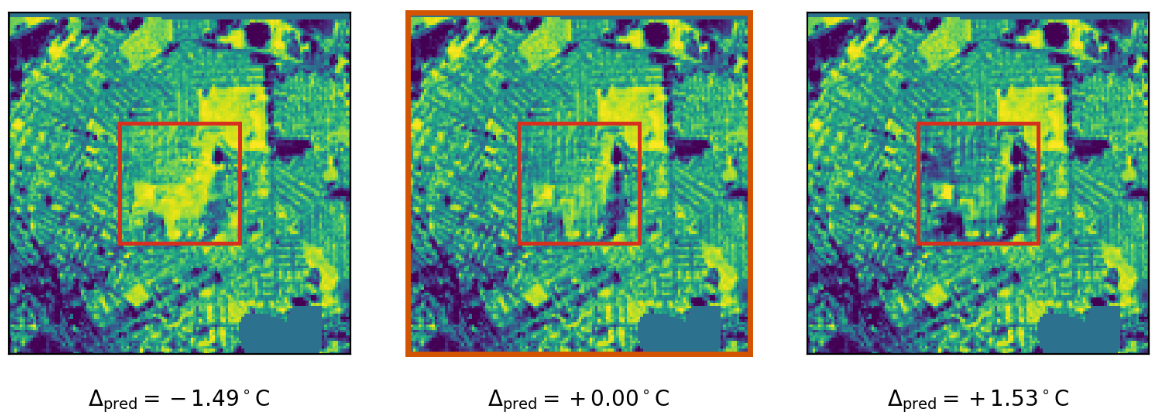}\vspace{4pt}

    \caption*{Figure~\ref{fig:supp_city_rows} (continued).}
\end{figure*}

\subsection*{S3. Implementation Details}
\paragraph{Forward model implementation details.}
The forward predictor $g_\phi$ is implemented as a U-Net using \texttt{segmentation\_models\_pytorch} with a ResNet-50 encoder and ImageNet-pretrained encoder weights. We train the model for 15 epochs with batch size 32 using Adam, a learning rate of $3\times 10^{-4}$, weight decay $10^{-4}$, and an MSE loss. During training, we apply an urban-weighted per-pixel loss with weight 5.0 on urban pixels, where the urban mask is used only for loss weighting. Urban pixels are defined as locations with positive building-height values. The best checkpoint is selected based on validation MAE on urban pixels, and the trained forward model is then frozen during inverse model training.

\paragraph{End-to-end inverse U-Net baseline.}
The end-to-end inverse U-Net baseline uses the same \texttt{segmentation\_models\_pytorch} U-Net architecture as the forward predictor, with a ResNet-50 encoder and ImageNet-pretrained encoder weights. The model is trained for 20 epochs with batch size 32 using Adam, a learning rate of $3\times10^{-4}$, no weight decay, and an L1 loss. The best checkpoint is selected based on NDVI MAE on the validation set, with the best model obtained at epoch 19.

\paragraph{Inverse diffusion model implementation details.}
The full architecture, optimization, training, and sampling hyperparameters are summarized in Table~\ref{tab:inv_diffusion_hparams}.

\begin{table}[h!]
\centering
\caption{Inverse diffusion model hyperparameters.}
\label{tab:inv_diffusion_hparams}
\begin{tabular}{ll}
\toprule
Parameter & Value \\
\midrule
\multicolumn{2}{l}{\textbf{Architecture}} \\
Backbone & NCSN++ \\
Image size & $128 \times 128$ \\
Base channels ($nf$) & 128 \\
Channel multipliers & $(1,2,2,2)$ \\
Residual blocks / resolution & 4 \\
Attention resolutions & $(16)$ \\
Normalization & GroupNorm \\
Nonlinearity & swish \\
Residual block type & BigGAN \\
\addlinespace

\multicolumn{2}{l}{\textbf{Optimization}} \\
Batch size & 16 \\
Optimizer & Adam \\
Learning rate & $2\times10^{-4}$ \\
Adam $\beta_1$ & 0.9 \\
Gradient clipping & 1.0 \\
LR warmup steps & 5000 \\
\addlinespace

\multicolumn{2}{l}{\textbf{EDM}} \\
$\sigma_{\mathrm{data}}$ & 0.5 \\
$P_{\mathrm{mean}}$ & -1.2 \\
$P_{\mathrm{std}}$ & 1.2 \\
\addlinespace

\multicolumn{2}{l}{\textbf{Physics loss}} \\
Physics loss & L1 \\
Physics warmup steps & 5000 \\
Physics ramp steps & 10000 \\
\addlinespace

\multicolumn{2}{l}{\textbf{Sampling}} \\
Sampling method & EDM \\
Sampler steps & 40 \\
$\rho$ & 7.0 \\
$\sigma_{\min}$ & 0.002 \\
$\sigma_{\max}$ & 80.0 \\
\bottomrule
\end{tabular}
\end{table}

\end{document}